\pdfoutput=1

\documentclass[11pt]{article}

\usepackage{acl}

\usepackage{times}
\usepackage{latexsym}

\usepackage[T1]{fontenc}

\usepackage[utf8]{inputenc}

\usepackage{microtype}
\usepackage{amssymb}
\usepackage{amsopn}
\usepackage{graphicx}
\usepackage{multirow}
\usepackage{makecell}
\usepackage[english]{babel}
\usepackage{bm}
\usepackage{array}
\usepackage{setspace}
\usepackage{CJKutf8}
\usepackage{algorithmicx}  
\usepackage{algpseudocode}  
\usepackage{amsmath}  
\usepackage{color,xcolor}
\usepackage{algorithm} 

\usepackage{pifont}
\usepackage{bm}
\usepackage{amsfonts}

\usepackage{graphicx}
\usepackage[normalem]{ulem}
\usepackage{multirow}
\usepackage{amsmath}
\usepackage{url}

\usepackage{tikz}
\usepackage{xcolor}
\usepackage{tcolorbox}
\usepackage{color,xcolor}

\usepackage{amssymb}
\usepackage{amsopn}
\usepackage{graphicx}
\usepackage{multirow}
\usepackage[english]{babel}
\usepackage{bm}
\usepackage{array}
\usepackage{setspace}
\usepackage{todonotes}

\usepackage{xspace}
\usepackage{amsfonts}
\usepackage{array,multirow}
\usepackage{pgfplots}
\usepackage{tikz}
\usepackage{verbatim}
\usepackage{arydshln}
\usepackage{booktabs}
\definecolor{ugreen}{rgb}{0,0.5,0}
\definecolor{mygreen}{RGB}{58,127,88}
\definecolor{iyellow}{RGB}{255,250,205}
\definecolor{ipurple}{RGB}{230,230,250}
\definecolor{myred}{RGB}{160,52,52} 
\definecolor{myblue}{RGB}{30,144,255}
\definecolor{myorange}{RGB}{255,127,80}
\definecolor{mypurple}{RGB}{255,20,147}

\title{Soft Language Clustering for Multilingual Model Pre-training}

\author{Jiali Zeng$^{1}\thanks{\ \ Corresponding author.}$, \ Yufan Jiang$^{3}$, \  Yongjing Yin$^{2}$, \ Yi Jing$^{4}$, \ Fandong Meng$^{1}$, \\ {\bf Binghuai Lin}$^{3}$, \ {\bf Yunbo Cao}$^{3}$, \ {\bf Jie Zhou}$^{1}$ \\
$^{1}$Pattern Recognition Center, WeChat AI, Tencent Inc \\
$^{2}$Westlake University $^{3}$Tencent Cloud Xiaowei $^{4}$Northeastern University\\
{\tt \{lemonzeng,fandongmeng,withtomzhou\}@tencent.com} \\
{\tt yinyongjing@westlake.edu.cn} \\ 
}

\begin{document}
\maketitle
\begin{abstract}

Multilingual pre-trained language models have demonstrated impressive (zero-shot) cross-lingual transfer abilities,
however, their performance is hindered when the target language has distant typology from source languages or when pre-training data is limited in size.
In this paper, we propose XLM-P, which contextually retrieves prompts as flexible guidance for encoding instances conditionally.
Our 
XLM-P enables (1) 
lightweight modeling of language-invariant and language-specific knowledge across languages, 
and (2) easy integration with other multilingual pre-training methods.
On the tasks of XTREME including text classification, sequence labeling, question answering, and sentence retrieval, both base- and large-size language models pre-trained with our proposed method exhibit consistent performance improvement. 
Furthermore, it provides substantial advantages for low-resource languages in unsupervised sentence retrieval and for target languages that differ greatly from the source language in cross-lingual transfer\footnote{Code and model are available at https://github.com/lemon0830/XLMP.git}.

\end{abstract}

\section{Introduction}

Multilingual pre-trained language models (mPLMs) such as mBERT \cite{Devlin-etal-2019-BERT}, mBART \cite{liu-etal-2020-multilingual-denoising}, XLM-R \cite{conneau-etal-2020-unsupervised} and mT5 \cite{xue-etal-2021-mt5} have lately produced notable advancements in a number of downstream NLP tasks.
In particular, the use of mPLMs significantly enhances few-shot fine-tuning and makes possible efficient zero-shot cross-lingual transfer \cite{Hu-etal-2020-XTREME}.
Essentially, an ideal mPLM should satisfy two properties: alignment between language pairs, which has been widely studied in the literature \cite{Chi-etal-2022-XLME, Ouyang-etal-2021-ERNIEM, chi-etal-2021-infoxlm}; and a good trade-off between high-resource and low-resource languages, which remains largely unexplored despite the success of mPLMs.

In this paper, we focus on the second property, specially the potential for model performance to suffer when a large number of languages are added.
This can occur due to restricted model capacity or computational limitations, resulting in underrepresented languages being allocated less capacity \cite{conneau-etal-2020-unsupervised}.
Furthermore, the model's coverage of world's languages remains inadequate, limiting the range of language technology applications it can support \cite{ansell-etal-2021-mad-g}.
A typical solution for the coverage-performance trade-off in multilingual learning is to assign additional model parameters to specific languages, such as language identity embeddings \cite{Conneau-etal-2019-xlm}, adaptors \cite{Houlsby-etal-2019-parameter,Ahmet-etal-2022-hyperX,ansell-etal-2021-mad-g}, and language-aware layers \cite{zhang-etal-2021-share}.
However, it is impractical for multilingual pre-training to maintain a separate component for each language, which can lead to more complicated and challenging optimization, especially for low-resource languages.

We propose to approach 
the above language-aware components from a different perspective.
In linguistic typology, some patterns such as nominative-accusative alignment have broad global distributions, whereas others like morphology are more specific and detailed \cite{donohue2008typology}.
To take advantage of this, we introduce XLM-P, which uses a set of compact embeddings to represent soft clustering of the language patterns beyond language identity.
We refer these embeddings as \textit{prompts}, due to their similarity to prompt tuning \cite{lester-etal-2021-power}.
Concretely, we build a key-value prompt pool and use the attention mechanism to look up the prompts for each input.
The retrieved prompts are then prepended to the input embeddings, and serve as categorization information to adapt the model weights conditionally.
This allows for more efficient and effective multilingual learning by leveraging the patterns and similarities across languages rather than maintaining separate components for each language.

We evaluate the proposed XLM-P on {\it Cross-Lingual Natural Language Understanding} tasks and {\it Cross-Lingual Sentence Retrieval} tasks of the XTREME benchmark, and the consistent improvement in performance demonstrates its effectiveness.
In addition, we conduct empirical analyses to investigate the underlying reasons of the improvement of XLM-P.
The advantages of XLM-P can be summed up as follows:
\begin{itemize}
    \item The prompt pool and instance-wise prompt retrieval are lightweight and only result in 0.35\% and 0.23\% increase in parameters for the base and large models, respectively.
    When fine-tuning on downstream tasks, the prompt module can be easily added or removed as needed.
    \item Our XLM-P divides the prompts into general and specific ones without any explicit supervision. The dynamically retrieved instance-wise prompts tame the sentence encoding, thus enhancing the capability of multilingual pre-trained models.
    \item
    The prompt module is model-agnostic and can be outfitted with the other frameworks (e.g., encoder-decoder style PLMs) and multilingual pre-training objectives (e.g., contrastive learning used in this paper).
    
\end{itemize}
Overall, XLM-P is a versatile and efficient approach for improving multilingual pre-training.

\section{Related Work}

\subsection{Cross-lingual LM pre-training}
Trained by the masked language modeling (MLM) loss with a shared multilingual vocabulary, multilingual BERT \cite{Devlin-etal-2019-BERT} achieves promising results in cross-lingual natural language understanding tasks \cite{Hu-etal-2020-XTREME}, which has attracted increasing attention to improve the cross-lingual transferability.
XLM-R \cite{conneau-etal-2020-unsupervised} increase the model capacity and use large-scale monolingual training data. 
In addition to monolingual data, XLM \cite{Conneau-etal-2019-xlm} performs MLM on bilingual parallel corpus, while ALM \cite{Yang-etal-2020-ALM} constructs code-switched sequences.
In respect of training objectives, a series of studies have explored various pre-training tasks to enhance the models' transferability \cite{huang-etal-2019-unicoder,Ouyang-etal-2021-ERNIEM,chi-etal-2021-improving,Chi-etal-2022-XLME, chi-etal-2021-infoxlm,luo-etal-2021-veco}. 

Compared to the above methods, our XLM-P exploits a small number of compact prompt vectors to tame the sentence encoding.
Moreover, the prompt module is model-agnostic, and can be combined with the above methods to achieve further improvement.

\subsection{Language-aware Components}
To alleviate the issue of the ‘curse of multilinguality’, various language-aware components have been proposed, which can allocate additional capacity for individual languages especially under-represented languages.
\citet{Conneau-etal-2019-xlm} use language identity embeddings to explicitly guide the model.
\citet{ansell-etal-2021-mad-g} present MAD-G, which contextually generates language adapters from language representations.
\citet{Ahmet-etal-2022-hyperX} propose a single hypernet-work that unifies multi-task and multilingual learning with efficient adaptation.
On multilingual neural machine translation, 
\citet{philip-etal-2020-monolingual} trained language-specific adapters.
\citet{zhang-etal-2021-share} use conditional routing to select shared and language-specific parameters.
\citet{stickland-etal-2021-recipes} use language-agnostic task adapters for fine-tuning BART and mBART to bilingual and multilingual MT.

We differ from them in that we do not use any language indicators, and regard the prompts as a bottleneck for storing a small number of discriminative features.
The prompts are encoded by the large language model along with the input, which learns language-invariant and language-specific features via the deep modular interaction.

\subsection{Prompt-based Tuning}
Our work builds upon the recent results showing the effectiveness of adapting PLMs to downstream tasks conditioning on lightweight prompt vectors \cite{Brown-etal-2020-languagemodelasfewshotlearner, lester-etal-2021-power, schick-schutze-2021-exploiting, Sanh-etal-2021-multiprompt}.
Differentiable prompts \cite{li-liang-2021-prefix,lester-etal-2021-power,gu-etal-2022-ppt} show the power of adapting pre-trained language model to multiple downstream tasks by simply prepending a few learnable parameters to the input.
More recently, the effectiveness of prompting has been investigated in multilingual (or cross-lingual) tasks, which is largely unexplored despite the success of prompting in English \cite{zhao-schutze-2021-discrete, zhou2022enhancing, huang2022zero}.

Instead of exploring prompting in fine-tuning, we propose to adopt dynamic retrieval of prompts, which jointly optimized with the mPLM, as a method of soft language clustering to enhance multilingual pre-training. 

\section{Method}

\begin{figure}[!t]
\centering
\includegraphics[width=1.0\linewidth]{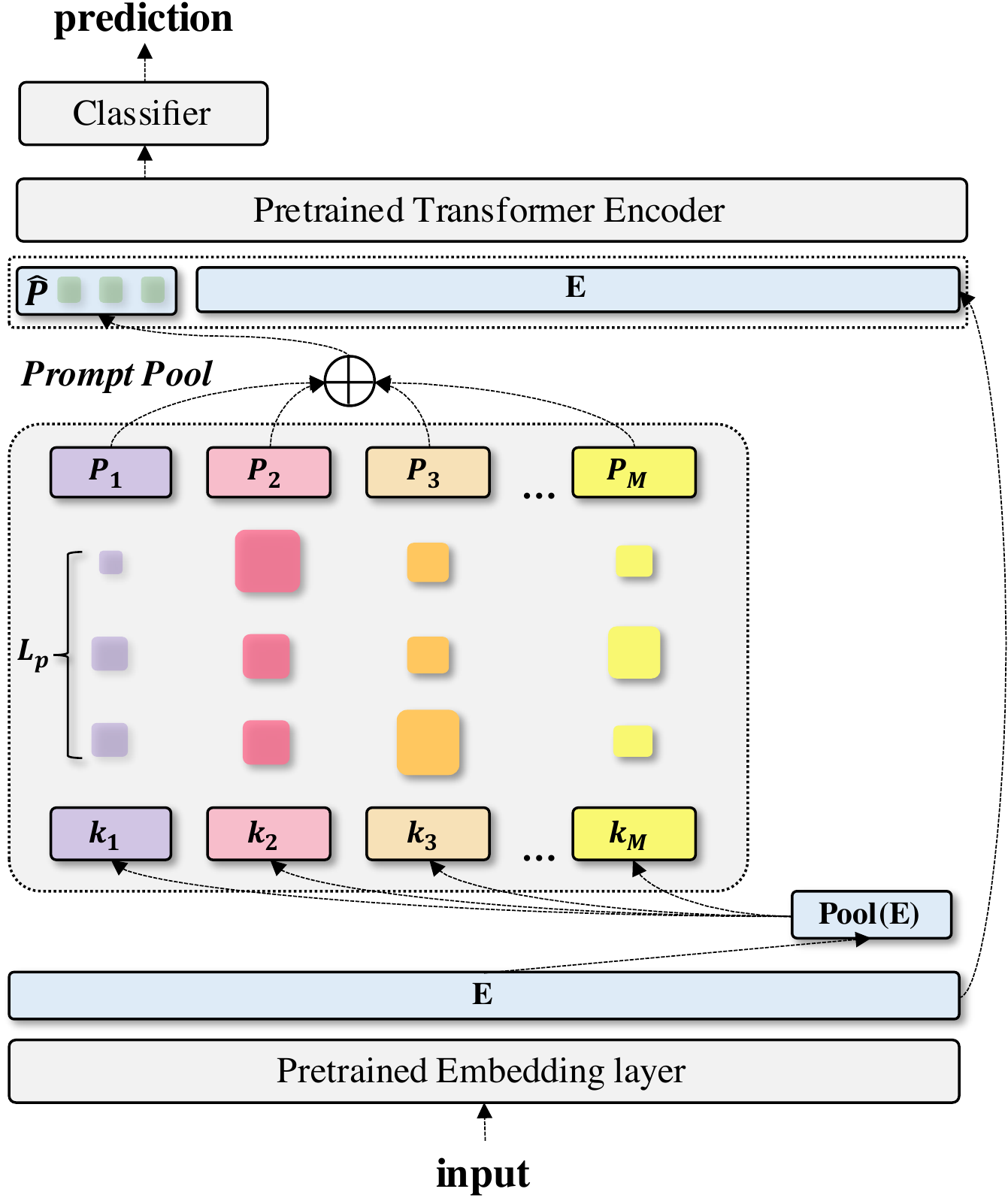}
\caption{
{\bf Model architecture of XLM-P}.
{
For each input, we look up the prompts in the key-value prompt pool \{($k_j$, $P_j$)\}, where $P_j \in \mathbb{R}^{L_p \times D}$ and $L_p$ is the number of vectors for each prompt.
The retrieved prompts are prepended to the input embeddings $E$, and serve as soft categorization information across languages}.
}
\label{fig_model}
\end{figure}

Figure \ref{fig_model} depicts the addition of a \textit{Prompt Pool} to the transformer-based language model.
Before feeding an input to the model, we perform \textit{Instance-wise Prompt Retrieval} and convert the input to a prompt-wrapped one.
Both the prompts and the model are jointly optimized on multilingual data using Masked Language Modeling,
which trains the model to make predictions based on both context and clustering information.
At the fine-tuning stage, we experiment with two strategies: standard fine-tuning and prompt-based fine-tuning.


\subsection{Prompt Pool}
In our proposed framework, 
we use a prompt pool to store fine-grained patterns sharing across languages as well as language-specific knowledge. 
Formally, the prompt pool is defined as:
\begin{equation}
    \textbf{P}=\{P_1, P_2, ..., P_M\},
\end{equation}
where $M$ is the total number of prompts in the prompt pool, {and $P_j \in \mathbb{R}^{L_p \times D}$ is a single prompt with $L_p$ vectors, whose dimension is the same as the embedding size $D$ of the mPLM}.
We associate each prompt as value to a learnable key:
$\{(k_1, P_1), (k_2, P_2), ..., (k_M, P_M)\}$, where $k_i \in \mathbb{R}^{D_k}$.
And we denote the set of all keys \textbf{K}=\{$k_j$\}$^M_{j=1}$.

\subsection{Instance-wise Prompt Retrieval}
Ideally, we would like to let the input itself to decide which prompts to choose through query-key matching.
Formally, given an input sentence \textbf{X}=$\{x_0, ..., x_n\}$, {we first map each token $x_i$ to a real-valued vector $e_i$ by an embedding layer.
The sequence of embeddings is then concatenated with the embedding of ``[CLS]'' token \textbf{E}=$\{e_{cls}, e_0, ..., e_n\}$.}
Next, we operate a pooling strategy\footnote{In this paper, we use \textit{max} pooling.} on the embedding sequence to obtain the input representation:
\begin{equation}
    r = {\rm Pool}(\textbf{E}).
\end{equation}
Next, we obtain the prompts for the input using the attention mechanism:
\begin{align}
    \hat{P} &= \sum \alpha_j P_j, \\
    \alpha_j &= {\rm} {\rm softmax}((rW)k_j^T),
\end{align}
where $W$ is a trainable parameter.
Finally, we prepend the retrieved prompt $\hat{P}$ to the input embeddings $\hat{\bf E}$=[$\hat{P}$, ${\bf E}$], and feed the extended sequence $\hat{\bf E}$ to the model to get the contextualized sentence representations ${\bf H}$=$\{h_{P_0}, .., h_{P_{L_p}}, h_{cls}, h0, ..., h_n\}$.

Notably, unlike language identity embeddings used by \citet{Conneau-etal-2019-xlm}, this soft manner allows the model to capture fine-grained 
language-invariant and language-specific features from massive monolingual data without explicitly assigning a cue to each input.

\subsection{Prompt-based Masked Language Modeling for Pre-training}

Our XLM-P is trained to predict the masked tokens using the prompt-augmented contextualized representations.
Following \citet{Devlin-etal-2019-BERT}, we randomly mask 15\% of the tokens in a monolingual sentence.
With probabilities of 0.8, 0.1, and 0.1, we replace each masked token with a special token ``[MASK]'', a random token, or the unchanged token, respectively.
The monolingual MLM loss is defined as:
\begin{equation}
   \mathcal{L}_{\rm MLM} = -\sum_{j \in M_x} {\rm log} p(x_i|{\bf H}, {\bf X}_{\setminus M_x})
\end{equation}
where ${\bf X}_{\setminus M_x}$ is the masked version of input ${\bf X}$.
Since the prompt module is agnostic to training objectives, we can enhance the model with additional objectives, as demonstrated in Section \ref{sec_main_Result}.

\subsection{XLM-P for Downstream Applications}

Benefiting from the plug-and-play property of the proposed prompt module, we can choose either plug in or plug out it on demand.
We explore two fine-tuning strategies to use our XLM-P on downstream applications, i.e., {\it Standard Fine-tuning} and {\it Prompt-based Fine-tuning}, and the detail is presented as follows.

\paragraph{Standard Fine-tuning.}
In this setting, we unplug the prompt module from the pre-trained model.
Concretely, we simply feed the embedding features ${\bf E}=\{e_{cls},e_0, ..., e_n\}$ excluding the prompt embeddings to the model, which is identical to XLM-R.

\paragraph{Prompt-based Fine-tuning.} 
We can keep the prompt module and conduct the retrieval during fine-tuning, as we do in training.
For token-level prediction tasks (e.g., structured prediction and question answering), we remove the hidden states of the prompt after encoding and take the rest ${\bf H}_{\setminus \hat{P}}$=$\{h_{cls}, h1, ..., h_n\}$ as the input of task-specific classifiers.
For sentence-level classification tasks, we apply a pooling operation on all of the hidden states 
${\bf H}_{\setminus X}$=$\{h_{P_0}, .., h_{P_{L_p}}, h_{cls}\}$ to obtain the sentence-level representation used for classification.

\begin{table*}[!t]
\centering
\small
\begin{spacing}{1.2}
\begin{tabular}{lccccccc}
\toprule
\multirow{2}{*}{\bf Task} &
 \multicolumn{2}{c}{\bf Structured Prediction} & \multicolumn{3}{c}{\bf Question Answering} & \multicolumn{2}{c}{\bf Classification} \\ 
& {POS} & {NER} & {XQuAD} & {MLQA} & {TyDiQA} & {XNLI} & {PAWS-X} \\
{\bf \#Languages} & 33 & 40 & 11 & 7 & 9 & 15 & 7 \\
{\bf Metrics} & F1 & F1 & F1/EM & F1/EM & F1/EM & Acc & Acc \\
\midrule
MBERT$_{base}$ {\dag} & 70.3 & 62.2 & 64.5/49.4 & 61.4/44.2 & 59.7/43.9 & 65.4 & 81.9 \\ 
XLM$_{base}$ {\dag} & 71.3 & 61.2 & 59.8/44.3 & 48.5/32.6 & 43.6/29.1 & 69.1 & 80.9 \\
MT5$_{base}$ {\dag} & - & 55.7 & 67.0/49.0 & 64.6/45.0 & 57.2/41.2 & 75.4 & 86.4 \\ 
XLM-E (-TRTD)$_{base}$ {\dag} & 74.2 & 62.7 & 74.3/58.2 & 67.8/49.7 & 57.8/40.6 & 75.1 & 87.1 \\ 
VECO$_{large}$ {\dag} & 75.1 & 65.7 & 77.3/61.8 & 71.7/53.2 & 67.6/49.1 & 79.9 & 88.7 \\ 
InfoXLM$_{large}$ {\dag} & - & - & - & 73.6/55.2 & - & {\bf 81.4} & -  \\
\midrule
\multicolumn{8}{l}{\textit{Backbones}} \\
XLM-R$_{base}$ {\dag}  & {75.6} & 61.8 & 71.9/56.4 & 65.1/47.2 & 55.4/38.3 & 75.0 & 84.9 \\ 
XLM-R$_{large}$ {\dag} & 72.6 & 65.4 & 76.6/60.8 & 71.6/53.2 & 65.1/45.0 & 79.2 & 86.4 \\ 
\multicolumn{8}{l}{\textit{Standard Fine-tuning}} \\
{\bf XLM-P$_{base}$} & 74.3 & {63.8} & 75.3/60.4 & 67.4/49.4 & 58.5/41.6 & 75.4 & 86.6 \\ 
{\bf XLM-P$_{large}$} & 76.9 & 68.1 & 79.0/63.9 & 72.4/53.5 & 72.1/55.0 & 81.1 & 88.7 \\ 
\multicolumn{8}{l}{\textit{Prompt-based Fine-tuning}} \\
{\bf XLM-P$_{base}$} & 73.9 & 63.6 & {75.8/61.5} & {68.7/50.1} & 59.3/42.8 & 75.1 & 86.0 \\ 
{\bf XLM-P$_{large}$} & {\bf 77.0} & {\bf 68.5} & {\bf 79.2/64.4} & {\bf 73.7/56.4} & {\bf 72.7/55.7} & {81.2} & {\bf 88.9} \\ 
\bottomrule
\end{tabular}
\end{spacing}
\caption{
\label{tab_results_main_result}
{\bf Experimental results on XTREME cross-lingual natural language understanding tasks}.
The methods with {\dag} denote that we directly report scores from the corresponding work.
We run 5 times with different random seeds and report the averaged results.
}
\end{table*}


\section{Experiments}

\subsection{Settings}
\paragraph{Pre-training.}
To train XLM-P, we extract a subset from CC-100 \cite{conneau-etal-2020-unsupervised} which involves monolingual data in 50 languages.
We use XLM-R \cite{conneau-etal-2020-unsupervised} as the backbone.
The XLM-P$_{base}$ model has 12 layers with 768 hidden units and 12 attention heads, and the XLM-P$_{large}$ model has 24 layers with 1024 hidden units and 16 attention heads.
We set $M$=256 and $L_p$=4 for both base model and large model, and XLM-P introduces 983,040 and 1,310,720 parameters to the original pre-trained model, merely accounting for 0.35\% and 0.23\% of the total parameters, respectively.
The detail of the pre-training settings can be found in Appendix \ref{Appendix_pretrain_setting}.

\paragraph{Evaluation.}
We evaluate our model on XTREME \cite{Hu-etal-2020-XTREME}, 
which is designed to assess 
the cross-lingual generalization capabilities of pre-trained language models, 
with a specific focus on {\it Cross-lingual Natural Language Understanding} and {\it Cross-lingual Sentence Retrieval}.
There are seven tasks for cross-lingual natural language understanding, which can be grouped into three categories:
1) Structured prediction: part-of-speech tagging (POS) on the Universal Dependencies v2.5 \cite{zeman-etal-2019-universal}, and named entity recognition (NER) on the WikiAnn \cite{pan-etal-2017-cross} dataset;
2) Question answering: cross-lingual question answering on MLQA \cite{lewis-etal-2020-mlqa} and XQuAD \cite{artetxe-etal-2020-cross}, and gold passage of typologically diverse question answering (TyDiQA-GoldP, \citet{clark-etal-2020-tydiqa});
3) Sentence classification: cross-lingual natural language inference (XNLI, \citet{conneau-etal-2018-xnli}), and cross-lingual paraphrase adversaries from word scrambling (PAWS-X, \citet{yang-etal-2019-paws}).
The aim of the cross-lingual sentence retrieval task is to retrieve relevant sentences across languages, and
we use the Tatoeba \cite{artetxe-schwenk-2019-massively} dataset.

\paragraph{Baselines.}
As baselines, we employ the multilingual pre-trained language models listed below:
1) MBERT \cite{Devlin-etal-2019-BERT} is pre-trained with MLM and next sentence prediction on Wikipedia in 104 languages;
2) XLM \cite{Conneau-etal-2019-xlm} is
pre-trained with MLM on 100 languages and translation language modeling (TLM) on 14 language pairs; 
3) MT5 \cite{xue-etal-2021-mt5} is the multilingual version of T5 pre-trained with text-to-text tasks;
4) XLM-E \cite{Chi-etal-2022-XLME} is trained with two pre-training tasks: namely multilingual replaced token detection (MRTD) and translation replaced token detection (TRTD).
We report XLM-E (-TRTD) for fair comparison, which does not use parallel data during pre-training;
and 5) VECO \cite{luo-etal-2021-veco} is a unified cross-lingual language model for both NLU and NLG.
6) InfoXLM \cite{chi-etal-2021-infoxlm} is jointly pre-trained with a cross-lingual contrastive learning task.

\subsection{Main Results} \label{sec_main_Result}

\paragraph{Cross-Lingual Natural Language Understanding.}

Following \citet{Hu-etal-2020-XTREME}, we adopt the zero-shot transfer setting for evaluation, in which the models are fine-tuned on English training data but evaluated on all the target languages.
Rather than selecting a single model for each language, we use only one model for evaluation.
The detail of the hyper-parameters used for fine-tuning can be found in Appendix \ref{Appendix_ft_setting}.

The results, which are averaged across all the target languages and five runs with different random seeds, are illustrated in Table \ref{tab_results_main_result}.
Compared to the XLM-R based models, the XLM-P based models achieve significantly better performances.
Besides, our XLM-P{$_{base}$} consistently outperforms the baselines MBERT{$_{base}$}, { MT5$_{base}$}, and {XLM-E (-TRTD)$_{base}$}, which are pre-trained
without any parallel corpora.
Moreover, {XLM-P$_{large}$} brings notable improvements over all the baselines on most of the tasks.
Concretely, XLM-P models perform better on the structure prediction and the question answering tasks, while preserving competitive results on the sentence classification tasks.
The overall experimental results demonstrate that multilingual pre-training can benefit from the guidance of our proposed prompt module.

Surprisingly, there appears to be minimal difference in performance between standard fine-tuning and prompt-based fine-tuning.
This can be explained by the fact that only English training data was used during fine-tuning, and the prompt embeddings were not specifically optimized for the task at hand.
Furthermore, recent studies have highlighted the challenges of prompt tuning in cross-lingual natural language understanding and have shown that this area is gaining more attention \cite{Qi-etal-2022-enhancing, huang-etal-2022-zero}.
We plan to explore this further in future work.
Finally, we compare our model with XLM-R under the translate-train-all setting and the results are reported in Appendix \ref{Appendix_translate}. These results further support the effectiveness of our model.


\paragraph{Cross-Lingual Sentence Retrieval.}
Following \citet{chi-etal-2021-infoxlm} and \citet{Hu-etal-2020-XTREME}, we use 14 and 36 languages of the parallel corpora for evaluation, respectively.
For sentence representations, we take the average of hidden states in a specific layer, and we use the 10-th layer for XLM-P.
Then, we induce translation pairs using a nearest neighbor search with cosine similarity.
As illustrated in Table \ref{tab_results_tatoeba}, XLM-P achieves 64.0 and 61.7 accuracy scores on Tatoeba-14, and 63.8 and 61.0 accuracy scores on Tatoeba-36 in the directions of en $\rightarrow$ xx and xx $\rightarrow$ en, respectively, which outperforms XLM-R significantly.
The improvement possibly due to that the sentence representations obtained from XLM-P encoding extra shared features across languages learned by our prompts. 
Moreover, under the setting of pre-training on multilingual monolingual corpus, {XLM-P} performs greatly better than {XLM-E (-TRTD)} and competes with {InfoXLM (-XLCO)}.

\begin{table}[!t]
\centering
\small
\begin{spacing}{1.1}
\setlength{\tabcolsep}{1.7mm}{
\begin{tabular}{lcccc}
\toprule
\multirow{2}{*}{\bf Model} & \multicolumn{2}{c}{\bf Tatoeba-14} & \multicolumn{2}{c}{\bf Tatoeba-36} \\
& en $\rightarrow$ xx & xx $\rightarrow$ en & en $\rightarrow$ xx & xx $\rightarrow$ en \\
\midrule
XLM-R {\dag} & 59.5 & 57.6 & 55.6 & 53.4 \\
XLM-E {\dag} & 74.4 & 72.3 & 65.0 & 62.3 \\
 \ \ -TRTD {\dag} & 55.8 & 55.1 & 46.4 & 44.6 \\
InfoXLM {\dag} & {\bf 80.6} & {\bf 77.8} & 68.6 & 67.3 \\
 \ \ -XLCO {\dag} & 64.6 & 65.3 & 50.9 & 53.5 \\
\midrule
{\bf XLM-P} & 64.0 & 61.7 & 63.8 & 61.0 \\
{\bf XLM-P+} & 73.2 & 77.2 & {\bf 76.4} & {\bf 69.0} \\
\bottomrule 
\end{tabular}}
\end{spacing}
\caption{
\label{tab_results_tatoeba}
{\bf{Experimental results on cross-lingual sentence retrieval task Tatoeba in terms of accuracy (\%)}}.
Methods with {\dag} denote that we directly report the scores from \cite{Chi-etal-2022-XLME}.
}
\end{table}

Notably, {InfoXLM} outperforms our {XLM-P} due to the benefit of the additional cross-lingual contrastive objective.
Our prompt module is designed to be compatible with other multilingual pre-training techniques, and to validate this, we added a simple dropout-based InfoNCE objective \cite{Oord-etal-2018-representation, He-etal-2020-momentumcl} to our XLM-P model.
More detailed is introduced in Appendix \ref{Appendix_drop_infonce}.
We post-train XLM-P on monolingual corpora in 50 languages with both MLM loss and InfoNCE loss.
The resulting model, XLM-P+, gives a significant improvement over {XLM-P} and performs better than XLM-E and InfoXLM on Tatoeba-36, but slightly worse on Tatoeba-14 against InfoXLM.
This is due to the fact that we did not use the parallel training data utilized by InfoXLM and XLM-E\footnote{
There are promising training strategies for pre-training mPLM using parallel corpus with our XLM-P,
such as the interaction between retrieved prompts of sentence pairs. 
This is an intriguing research question that merits further investigation in future studies.}. 
In a nutshell, the results show that our proposed method can be effectively integrated with other pre-training objectives.

\begin{figure}[!t]
\centering
\includegraphics[width=1.0\linewidth]{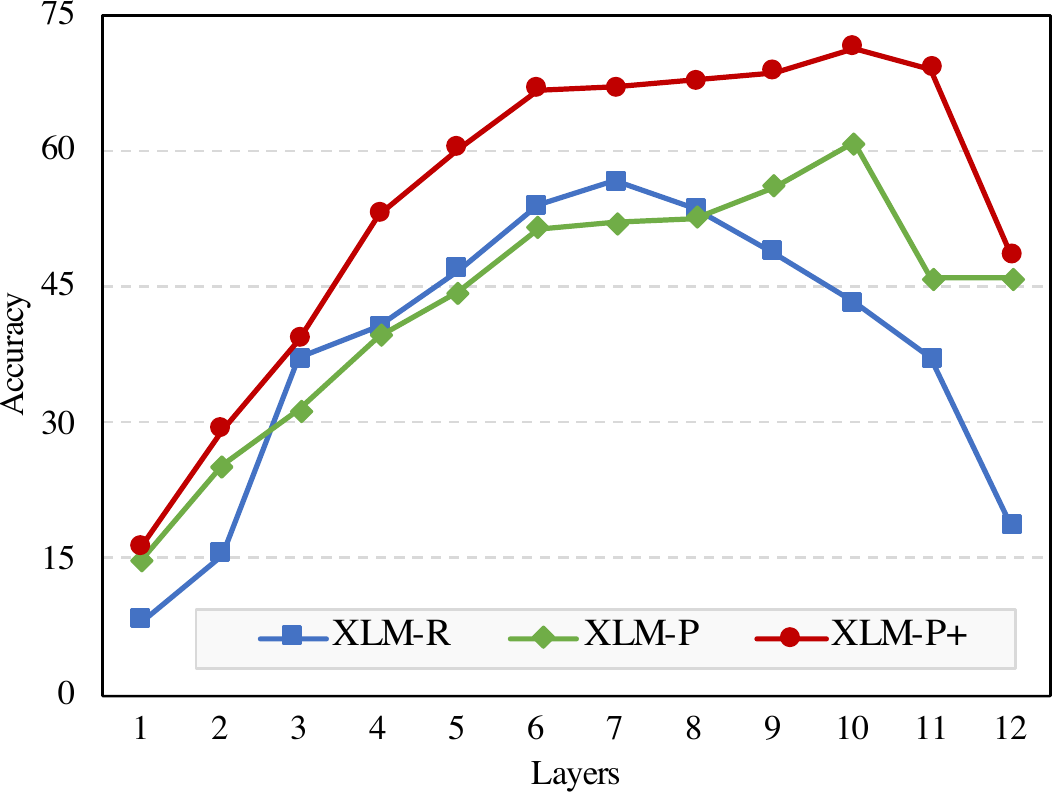}
\caption{
{\bf Evaluation results on Tatoeba cross-lingual sentence retrieval over different layers}.
For each layer, the accuracy score is averaged over all the 36 language pairs in the xx $\rightarrow$ en direction.
}
\label{fig_retrieval_layers}
\end{figure}

Finally, in Figure \ref{fig_retrieval_layers},
we illustrate the effect of layer selection on sentence representations.
The accuracy score is calculated by taking the average of all the 36 language pairs in xx $\rightarrow$ en directions.
The figure shows that all the models exhibit a parabolic trend across layers. 
Different from {\it XLM-R} that achieves the highest accuracy of 56.5 at the 7-th layer, the curve of {\it XLM-P} rises more steadily until it peaks at the 10-th layer.
It can be observed that {\it XLM-P} outperforms {\it XLM-R} on the top layers, and {\it XLM-P+} achieves notably higher average scores than {\it XLM-R} at all layers.

\section{Analysis}
We carry out a number of analyses in order to comprehend the design of our proposed XLM-P better.

\begin{figure}[!t]
\centering
\includegraphics[width=1.0\linewidth]{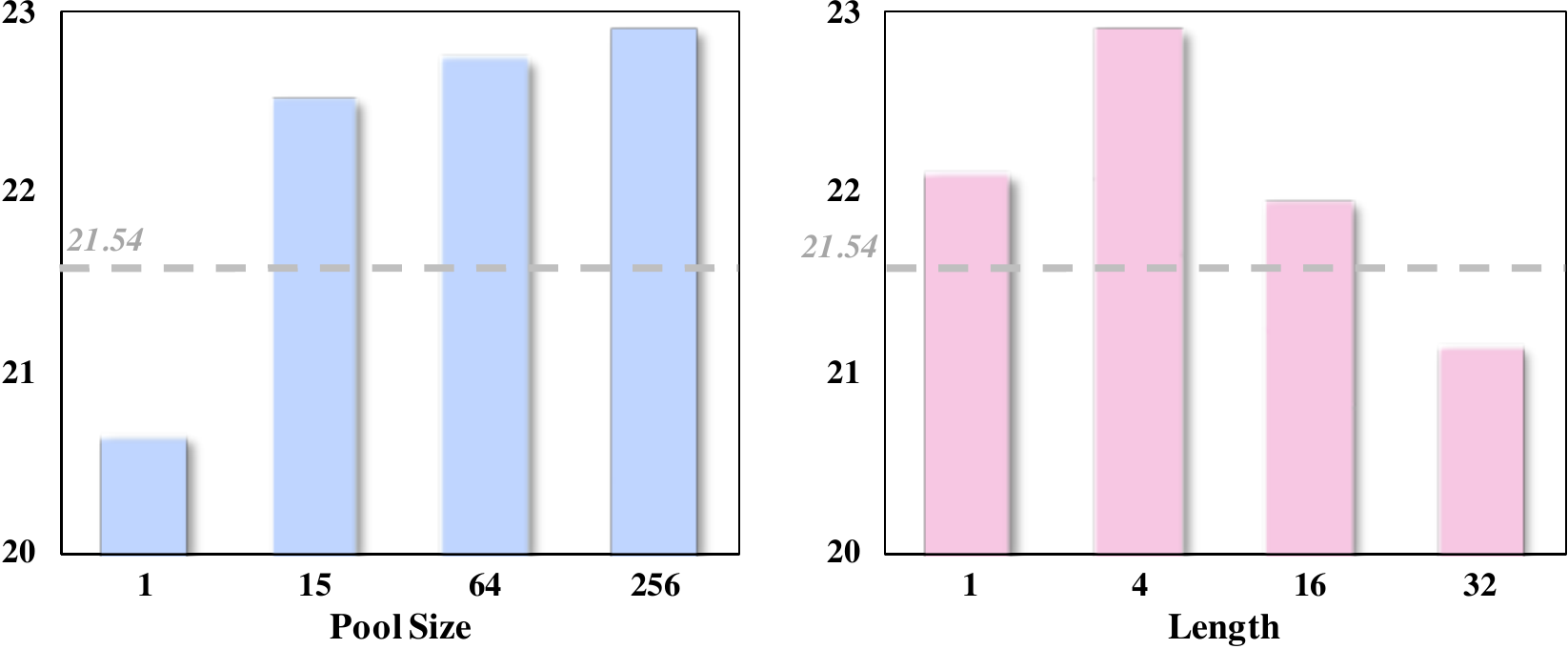}
\caption{
{\bf Effects of Prompt Capacity.}
Left: Average score w.r.t prompt pool size $M$ given $L_p=4$.
Right: Average score w.r.t. prompt length $L_p$ given $M=256$.
Dashed lines indicate the results of XLM-R$_{small}$ under standard fine-tuning.
}
\label{fig_hyperparameters}
\end{figure}

\begin{figure*}[!th]
\centering
\includegraphics[width=1.0\linewidth]{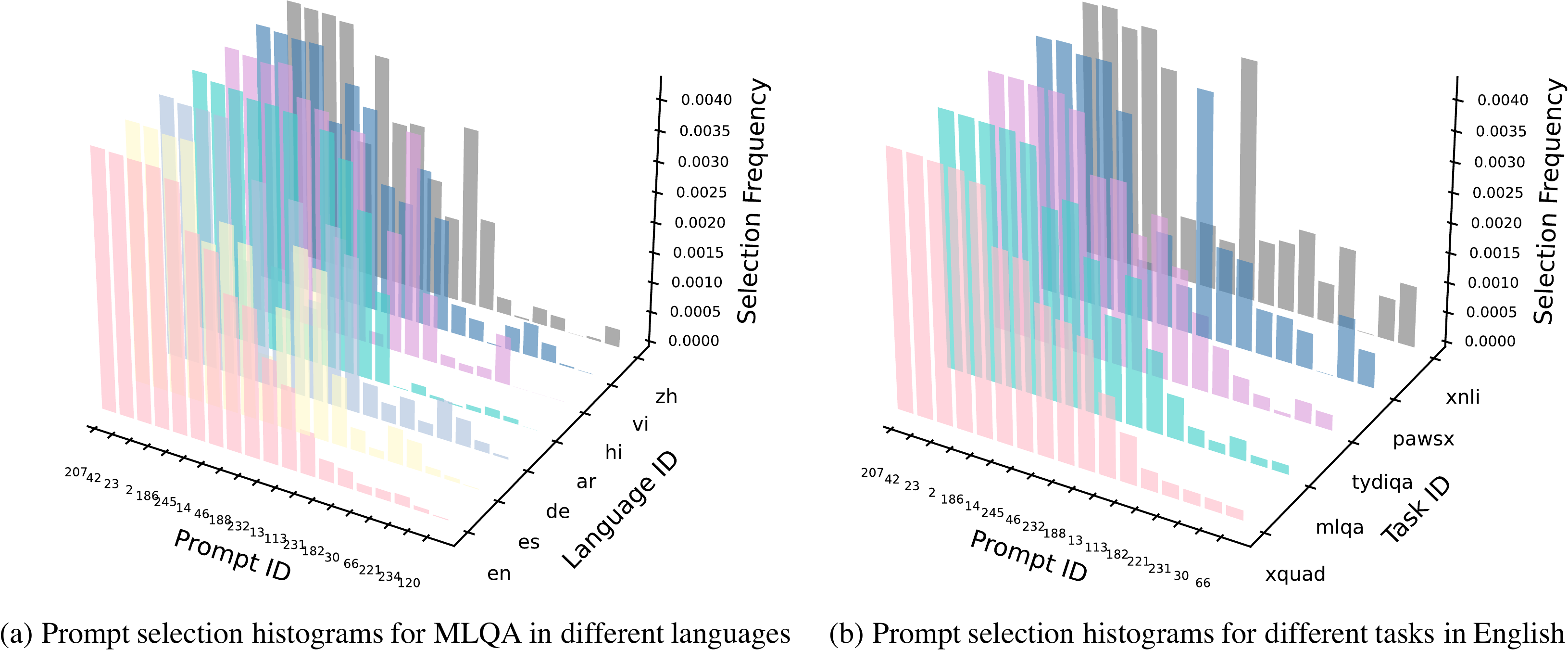}
\caption{
{\bf Prompt selection histograms}.
We only display the top 16 selected prompts.
{The prompts retrieved can be grouped into language-shared and -specific prompts. 
Similar tasks tend to have similar patterns of prompt selection across different datasets, whereas tasks that are more irrelevant to each other prefer selecting more diverse prompts.}
}
\label{fig_prompt_select}
\end{figure*}

\subsection{Effect of Prompt Capacity}
In XLM-P, the length of a simple prompt $L_p$ and the size of the prompt pool $M$ are two critical hyper-parameters that determine the total capacity of the learnable prompt embeddings.
$M$ determines the granularity of prompt selection and $L_p$ decides the expressiveness of a single prompt.
To investigate the effects of the two hyper-parameters, we train small-size XLM-R and XLM-P from scratch and keep the other setting unchanged
(see Appendix \ref{Appendix_pretrain_setting} for more details).
We evaluate the pre-trained models on TydiQA and report the average of F1 and EM scores in Figure \ref{fig_hyperparameters}.

We can see that removing the prompt pool and only using a single prompt (i.e., pool size $M=1$) results in a significant performance drop, suggesting that it is not sufficient to maintain a single shared prompt for all languages.
Increasing the size of prompt pool shows a positive effect on performance, but excessive prompts degrade the performance.
The result verifies our motivation, that the prompts are used to capture abstract clustering information.
Too many prompts can dilute the effect, and thus negatively affect the generalization and transferability of representations.

\begin{figure*}[!t]
\centering
\includegraphics[width=1.0\linewidth]{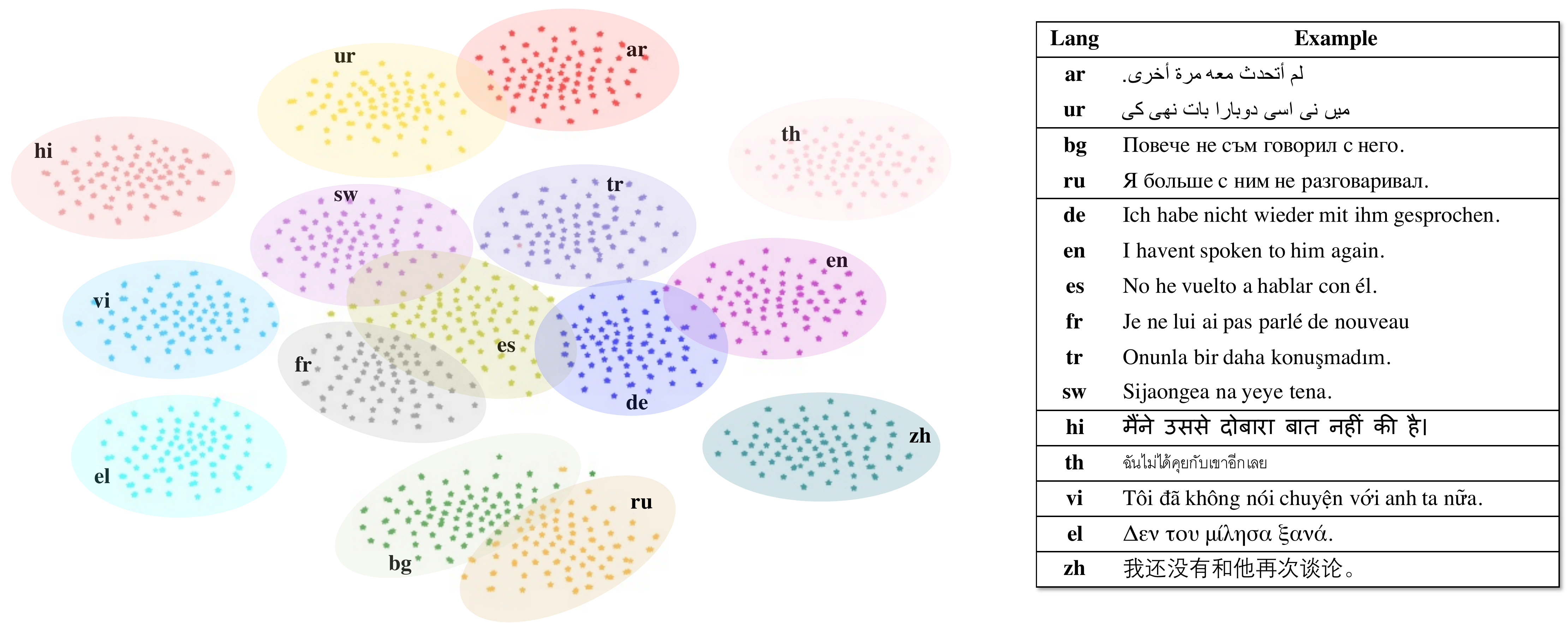}
\caption{
{\bf Visualization of prompts retrieved by sentences in different languages}.
We shows one sentence in each language for a clearer understanding.
The representations of prompts retrieved by sentences in various languages are distinct, while the representations of prompts retrieved by sentences in similar languages are more alike.
}
\label{fig_prompt_emb}
\end{figure*}

\subsection{Patterns of Retrieved Prompts} \label{sec_pattern_retrieved_prompt}
As depicted in the left portion of Figure \ref{fig_prompt_select},
we present the prompt selection histograms for XLM-P on test sets of MLQA \textit{in different languages}.
The prompts retrieved can be easily distinguished into two categories: language-shared prompts (e.g., prompt 207, 42, 23 and 2) and language-specific prompts (e.g., prompt 14, 66, 221, 120).
In addition, we display the histograms on test sets of \textit{different tasks} in English in the right portion of Figure \ref{fig_prompt_select}. 
In particular, 
the similar tasks tend to have similar patterns of prompt selection across different datasets (e.g., xquad and mlqa, both of which are question answering tasks), whereas the tasks that are more irrelevant to each other (e.g., xquad and pawsx, pawsx is a sentence classification task) prefer selecting more diverse prompts.
This phenomenon echos the effectiveness of our XLM-P.
Even without any explicit supervision, our model can learn to group the prompts into general and specific ones.

\begin{table*}[!t]
\centering
\small
\begin{spacing}{1.1}
\setlength{\tabcolsep}{1.1mm}{
\begin{tabular}{lccccccccccccccccc}
\toprule
\multirow{2}{*}{\bf Model} & & \multicolumn{7}{c}{\bf High Resource} & \multicolumn{7}{c}{\bf Low Resource} & \multirow{2}{*}{Avg.} & \multirow{2}{*}{Std.($\downarrow$)} \\
\cmidrule(lr){3-9}\cmidrule(lr){10-16}
& Direction & de & pt & nl & fr & es & ru & vi & eu & te & tl & bn & ta & sw & jv \\
\midrule
XLM-R {\dag} & xx $\rightarrow$ en & 89.9 & 80.6 & 79.5 & 74.1 & 74.0 & 72.5 & 68.4 & 33.5 & 32.5 & 31.2 & 29.3 & 25.7 & 18.7 & 15.1 & 51.79 & 27.04 \\
InfoXLM {\dag} & xx $\rightarrow$ en & {\bf 93.9} & 84.7 & 80.8 & 79.4 & {\bf 88.2} & {\bf 83.8} & {\bf 89.6} & 36.7 & 53.0 & 42.1 & {\bf 49.6} & {\bf 53.7} & {\bf 39.5} & 13.2 & 63.44 & 25.28 \\
{\bf XLM-P+} & xx $\rightarrow$ en & 93.3 & {\bf 87.3} & {\bf 88.9} & {\bf 82.8} & 87.4 & 82.0 & 82.8 & {\bf 60.4} & {\bf 58.5} & {\bf 59.1} & 47.8 & 49.2 & 39.3 & {\bf 35.2} & {\bf 68.14} & {\bf 20.28} \\
\midrule
XLM-R {\dag} & en $\rightarrow$ xx & 89.4 & 80.6 & 79.5 & 74.1 & 74.0 & 72.5 & 68.4 & 33.5 & 32.5 & 31.2 & 29.3 & 25.7 & 18.7 & 15.1 & 51.75 & 26.98 \\
InfoXLM {\dag} & en $\rightarrow$ xx & {\bf 95.1} & 86.5 & 81.8 & {\bf 84.0} & {\bf 87.2} & {\bf 85.7} & {\bf 92.0} & 28.6 & 53.0 & 35.5 & 49.1 & 63.5 & 40.8 & 7.8 & 63.61 & 27.85 \\
{\bf XLM-P+} & en $\rightarrow$ xx & 91.3 & {\bf 87.9} & {\bf 87.6} & 81.1 & 84.7 & 83.2 & 84.5 & {\bf 73.8} & {\bf 82.9} & {\bf 71.0} & {\bf 62.1} & {\bf 73.3} & {\bf 45.9} & {\bf 50.2} & {\bf 75.68} & {\bf 14.10}  \\
\bottomrule 
\end{tabular}}
\end{spacing}
\caption{
\label{tab_results_tatoeba_detail14}
{\bf{Experimental results on cross-lingual sentence retrieval task Tatoeba in terms of accuracy (\%)}}.
Methods with {\dag} denote that we directly report the scores from \cite{Chi-etal-2022-XLME}.
{XLM-P+} yields similar benefits across different data scales, and greatly benefits low-resource languages (e.g., eu, te, and jv).
}
\end{table*}

\subsection{Visualization of Retrieved Prompts}
Based on the observation in Section \ref{sec_pattern_retrieved_prompt}, we further plot t-SNE visualizations \cite{maaten2008sne} of prompts retrieved by sentences in 15 languages on XNLI test set\footnote{We sampled 200 sentences from each language.} in Figure \ref{fig_prompt_emb} to support our hypothesis that prompts can serve as a representation of soft clustering for language.

We can find that the prompt representations retrieved by the sentences in different languages are evidently distinct, while the prompt representations retrieved by the sentences in similar languages (e.g., ar and ur, bg and ru) are closer.
It implies that the prompts are able to capture the shared patterns across languages as well as language-specific patterns during multilingual pre-training.

\begin{table}[!t]
\centering
\small
\begin{spacing}{1.1}
\setlength{\tabcolsep}{1.0mm}{
\begin{tabular}{lccccccccc}
\toprule
\multirow{2}{*}{\bf Model} & Train & \multicolumn{3}{c}{\bf European Langs} & \multicolumn{3}{c}{\bf Asian Langs} & \multirow{2}{*}{Avg.} \\
 & {en} & {de} & {es} & {fr} & {ja} & {ko} & {zh} \\
\midrule
XLM-R &  87.8 & 94.3 & 88.6 & 88.8 & 77.1 & 75.1 & 80.4 & 84.59 \\
{\bf XLM-P} & {\bf 89.0} & {\bf 95.1} & {\bf 90.0} & {\bf 90.6} & {\bf 79.4} & {\bf 80.1} & {\bf 81.9} & {\bf 86.59} \\
\bottomrule 
\end{tabular}}
\end{spacing}
\caption{
\label{tab_results_detailed_pawsx}
{{\bf Experimental results on PAWS-X in terms of accuracy (\%)}. We adopt the zero-shot transfer setting for evaluation, in which the models are fine-tuned on English training data but evaluated on all the target languages.
}}
\end{table}

\subsection{Trade-off on High and Low Resource Languages}

In Section \ref{sec_main_Result}, we display the average performance across all of the languages, and we present more details about the performance in this section.
First, we choose the PAWS-X task with 7 languages including English (i.e., the language of the training set), three European languages (i.e., German, Spanish, French), and three Asian languages (i.e., Japanese, Korea, and Chinese).
As shown in Table \ref{tab_results_detailed_pawsx}, compared with XLM-R, 
XLM-P achieves better or more competitive performance on the test sets in European languages, while obtaining huge gains on the test sets in Asian languages, which are different from English.

Then,
we illustrate the results on Tatoeba in both the xx $\rightarrow$ en and en $\rightarrow$ xx directions in Table \ref{tab_results_tatoeba_detail14}. 
Due to the limited space, we only select the seven languages with the highest scores and the seven languages with the lowest scores based on the performance of XLM-R on Tatoeba-36 in xx $\rightarrow$ en direction, which can be grouped as {\it High} resource language and {\it Low} resource language pairs, respectively.
We compare our {XLM-P+} with {XLM-R} and {InfoXLM}, and report the mean and the standard deviation of accuracy.
Both {XLM-P+} and {InfoXLM} provide advantages in both directions.
Specifically, {XLM-P+} yields similar benefits across different data scales, and greatly benefits low-resource languages (e.g., eu, te, and jv).
By contrast, {InfoXLM} performs well on high-resource languages but has a marginal impact on low-resource languages.

We argue that mPLMs still suffer from {\it insufficient modeling capacity}, and adding more languages can result in a decline in representation quality.
Our proposed XLM-P can indeed alleviate this issue, especially for languages with limited data or languages that differ from the training data.
Intuitively, the dynamically retrieved instance-wise prompts in XLM-P make the sentence encoding specific to the soft clustering information, thereby enhancing the model capability of multilingual pre-trained models.

\section{Conclusion}
This paper presents XLM-P, a new multilingual pre-trained language model equipped with contextually retrieved prompts.
In particular, we prepend prompts-like learnable vectors to the input for modeling language interdependence and other potential sharing information.
Compared with other language-aware components, the retrieved prompts are parameter-efficient and more flexible without the requirement of language detection.
Experiments and analyses validate the effectiveness and robustness of XLM-P.
In addition, our method is compatible with various existing multilingual pre-training objectives.

\section*{Limitations}

In this paper, we simply prepend the retrieved prompt to the input embeddings before encoding.
A well-designed method of combining prompts with the input embeddings, such as Prefix Tuning \cite{li-liang-2021-prefix}, may result in additional enhancements.
Finally, as observed in Section \ref{sec_main_Result}, 
prompt-based fine-tuning does not present obvious superiority over standard fine-tuning.
Exploring the prompt tuning on cross-lingual natural language understanding is a challenging task that has recently gained attention \cite{Qi-etal-2022-enhancing, huang-etal-2022-zero}, and we leave it as future work.

\section*{Acknowledgements}
We would like to thank all of the anonymous reviewers for the helpful comments.

\bibliography{anthology,custom}
\bibliographystyle{acl_natbib}

\appendix

\begin{table}[!t]
\centering
\begin{tabular}{ccccc}
\toprule
\multicolumn{5}{c}{\bf Language} \\
af & {\bf ar} & {\bf bg} & bn & cs \\
{\bf de} & {\bf el} & {\bf en} & {\bf es} & et \\
eu & {\bf fr} & fa & fi & fy \\
gu & gd & he & {\bf hi} & hu \\
id & it & ja & jv & ka \\
kk & ko & lt & lv & ms \\
ml & my & mr & pl & pt \\
ne & nl & {\bf ru} & ro & si \\
{\bf sw} & ta & te & {\bf tr} & {\bf th} \\
tl & {\bf vi} & {\bf ur} & yo & {\bf zh} \\
\bottomrule
\end{tabular}
\caption{
\label{tab_language_code}
{\bf The language code of monolingual pre-training corpus}.
Among them, the bolded 15 languages are the languages included in the pre-training data of small-sized XLM-P and small-sized XLM-R.
}
\end{table}

\begin{table*}[!h]
\centering
\small
\begin{spacing}{1.2}
\begin{tabular}{lccc}
\toprule
{\bf Pre-training Hyperparameters} & {\bf Large} & {\bf Base} & {\bf Small} \\
\midrule
Number of layers & 24 & 12 & 4 \\
Hidden Size & 1024 & 768 & 768 \\
FFN inner hidden size & 4096 & 3072 & 3072 \\
Attention heads & 16 & 12 & 12 \\
Attention head size & 64 & 64 & 64 \\
Embedding Size & 1024 & 768 & 768 \\
Mask percent & 15\% & 15\% & 15\% \\
Warmup steps & 10k & 10k & 10k \\
Learning Rate & 1e-4 & 5e-4 & 3e-4 \\
Adam $\epsilon$ & 1e-6 & 1e-6 & 1e-6 \\
Adam $\beta_1$ & 0.9 & 0.9 & 0.9 \\
Adam $\beta_2$ & 0.98 & 0.98 & 0.98 \\
Attention Dropout & 0.1 & 0.1 & 0.1 \\
Dropout & 0.1 & 0.1 & 0.1 \\
Weight Decay & 0.01 & 0.01 & 0.01 \\
Max Sequence Length & 512 & 512 & 512 \\
Batch Size & 8,192 & 8,192  & 2,048 \\
Train Steps & 240k & 240k & 125k \\
Total Parameters & 561M & 279M  & 222M \\
\bottomrule 
\end{tabular}
\end{spacing}
\caption{
\label{tab_settings_pretraining}
{\bf The pre-training hyperparameters}.
}
\end{table*}

\section{Pre-training Details} \label{Appendix_pretrain_setting}
For monolingual data, following XLM-R \cite{conneau-etal-2020-unsupervised} and Veco \cite{luo-etal-2021-veco}, we build a clean CommonCrawl Corpus using an open-source tool CCNet \cite{Wenzek-etal-2020-ccnet}.
We use monolingual data in 50 languages for base-sized and large-sized XLM-P and monolingual data in 15 languages for small-sized XLM-R and XLM-P.
Table \ref{tab_language_code} reports the language codes for pre-training.
Please ref \cite{luo-etal-2021-veco} for the detailed data statistic of the monolingual pre-training corpus.
Following \citet{chi-etal-2021-infoxlm} and \citet{luo-etal-2021-veco}, we initialize the parameters of XLM-P with XLM-R \cite{conneau-etal-2020-unsupervised}.
We use the Adam optimizer \cite{Kingma-LeCun-2015-adam} with the learning rate 3e-4 for the base model and 1e-4 for the large model, respectively.
The full set of pre-training hyperparameters for small-sized, base-sized and large-sized XLM-P are listed in Table \ref{tab_settings_pretraining}.
We conduct the pre-training experiments using 64 Nvidia A100-40GB GPUs with 8,192 batch size for base and large XLM-P.


\begin{table*}[!h]
\centering
\small
\begin{spacing}{1.1}
\begin{tabular}{lccccccc}
\toprule
 & {\bf POS} & {\bf NER} & {\bf XQuAD} & {\bf MLQA} & {\bf TyDiQA} & {\bf XNLI} & {\bf PAWS-X} \\
\midrule
Batch size & \{8, 16, 32\} & 8 & 32 & 32 & 32 & 32 & 32 \\
Learning rate & \{1,2,3\}e-5 & \{5,..,9\}e-6 & \{2,3,4\}e-5 & \{2,3,4\}e-5 & \{2,3,4\}e-5 & \{5,...,8\}e-6 & \{10,20\}e-6 \\
Warmup & 10\%  & 10\%  & 10\%  & 10\%  & 10\%  & 12,500 steps & 10\% \\
Epochs & 10 & 10 & 4 & \{2,3,4\} & \{10,20,40\} & 10 & 10 \\
\bottomrule 
\end{tabular}
\end{spacing}
\caption{
\label{tab_settings_finetuning}
{\bf Hyperparameters used for fine-tuning on the XTREME end tasks}.
}
\end{table*}

\begin{table}[!t]
\centering
\small
\begin{spacing}{1.1}
\begin{tabular}{lcccc}
\toprule
{\bf Model} & NER & TyDiQA & PAWS-X & Avg. \\
\midrule
XLM-R$_{large}$ & 87.8$^{*}$ & 72.2/54.8 & 90.5 & 80.60 \\
{\bf XLM-P$_{large}$} & {\bf 91.1}$^{*}$ & {\bf 74.2/58.1} & {\bf 91.2} & {\bf 82.82} \\
\bottomrule 
\end{tabular}
\end{spacing}
\caption{
\label{tab_results_transall}
{\bf Experimental results of using pseudo-parallel data}.
Since \citet{Hu-etal-2020-XTREME} do not release the translation data of NER task, we use the golden NER training data of XTREME 
for reproduction. 
}
\end{table}

\section{Hyperparameters for Fine-tuning} \label{Appendix_ft_setting}
In Table \ref{tab_settings_finetuning}, we present the hyperparameters for fine-tuning baselines and our XLM-P on the XTREME end tasks.
For each task, the hyperparameters are searched on the joint validation set of all languages.

\section{Translate-Train-All Setting} \label{Appendix_translate}
In this section, we investigate another fine-tuning setting, {\it Translate-Train-All}, 
in which we fine tune a PLM on a mixed corpus consisting of golden training data in English and translated training data in other languages.

Table \ref{tab_results_transall} presents the results on NER, TyDiQA, and PAWS-X.
{XLM-P$_{large}$} outperforms {XLM-R$_{large}$} across all the tasks, confirming our model's effective capacity for cross-lingual transfer.

\section{Detailed of Dropout-based InfoNCE} \label{Appendix_drop_infonce}

Specifically, we construct the ``positive pairs'' by passing the same sentence to the model twice, and take other sentences in the same mini-batch as ``nagatives''.
We use the average of prompt hidden states and ``CLS'' hidden states of the latest layer as the sentence representation. 
The model is trained to predict the positive one among the samples as follows:
\begin{equation}
    l_i=-\log \frac{e^{{\rm sim}(v_{i}, v^{+}_{i})/\tau}}{\sum^{N}_{k=1} e^{{\rm sim}(v_{i}, v^{+}_{k})/\tau}}
\end{equation}
We post-train our XLM-P in monolingual corpora in 50 languages with the MLM loss and InfoNCE loss.
The learning rate is 5e-5, the total number of training step is 100k, and the warmup steps is 10k.

\end{document}